\documentclass{article}

\PassOptionsToPackage{numbers, compress}{natbib}
\usepackage[preprint]{neurips_2026}

\usepackage[utf8]{inputenc}
\usepackage[T1]{fontenc}
\usepackage[hidelinks]{hyperref}
\usepackage{url}
\usepackage{booktabs}
\usepackage{makecell}
\usepackage{multirow}
\usepackage{amsfonts}
\usepackage{amsmath}
\usepackage{amssymb}
\usepackage{nicefrac}
\usepackage{microtype}
\usepackage{graphicx}

\usepackage{array}        % needed by \newcolumntype
\usepackage{xcolor}
\usepackage{algorithm}
\usepackage{algorithmic}

\usepackage{adjustbox}

\newcommand{\sirenTableStyle}{%
  \small
  \setlength{\tabcolsep}{4pt}%
  \renewcommand{\arraystretch}{1.12}%
}

\graphicspath{{Figures/}}

\title{SIREN-Bench: Behavior-Driven Generation and Evaluation of Emergency-Vehicle Interactions}

\author{%
  Yicheng Zhu \\
  Rochester Institute of Technology \\
  \And
  Tianmu Zhao \\
  CVS Health \\
  \And
  Haoxin Leng \\
  City University of Hong Kong \\
  \And
  Fan Zuo \\
  New York University \\
  \And
  Tao Li \\
  City University of Hong Kong \\
  \And
  Zilin Bian \thanks{Corresponding author.} \\
  Rochester Institute of Technology \\
  \texttt{zilin.bian@rit.edu} \\
}

\begin{document}
\maketitle

% ============================================================
%  Abstract
% ============================================================
\begin{abstract}
Emergency vehicles (EMVs) can reorganize surrounding traffic as civilian vehicles brake, change lanes, or form rescue corridors in response to their passage. Evaluating these safety-critical interactions requires behavior-level control over both EMV privileges and civilian responses, together with consistent sensing and ground truth. Existing datasets and simulation benchmarks do not directly provide this combination. We present \textbf{SIREN}, a behavior-driven SUMO--CARLA co-simulation platform for generating EMV--civilian interactions. SIREN couples SUMO's network-level traffic evolution and behavior logic with CARLA's continuous vehicle control and synchronized onboard sensing; depending on the active behavior, the interaction is controlled by SUMO, CARLA, or jointly. We instantiate the platform as \textbf{SIREN-Bench-v1}, comprising seven parameterized interaction templates across emergency levels L1--L3 and three behavior families, with synchronized sensor observations and simulator-native annotations. We demonstrate the benchmark through three representative tasks: 3D object detection, trajectory prediction, and vision-language risk understanding. Evaluations of nine trajectory predictors, four LiDAR-based detectors, and five vision-language models reveal behavior-dependent failure modes. Traffic-clearance interactions are hardest for detection, privileged intersection traversal is hardest for prediction, and no learned predictor outperforms the constant-velocity reference on average. Vision-language models perform substantially better on normal traffic than on near-miss and collision events. These results demonstrate the value of behavior-centered benchmarking and establish SIREN as an extensible data-generation and evaluation platform for autonomous-driving and transportation safety research.
\end{abstract}

\providecommand{\keywords}[1]
{
  \small
  \textbf{\textit{Keywords---}} #1
}

% ============================================================
%  Paper body
% ============================================================
\section{Introduction}
Emergency vehicles (EMVs) do not merely add a rare vehicle type to ordinary traffic, instead, their passage can reorganize the behavior of the surrounding traffic. While responding to an emergency, an ambulance, fire truck, or police vehicle may cross a red signal, proceed through a stop-controlled intersection, pass between queued lanes, or temporarily occupy nonstandard road space. Civilian vehicles must react by braking, changing lanes, pulling toward lane boundaries, or entering otherwise restricted space to yield. These responses can extend beyond the vehicles directly blocking the EMV, altering lane occupancy, vehicle spacing, and right-of-way interactions across the local traffic scene \cite{clawson_wake-effectemergency_1997,hsiao_preventing_2018}. An EMV encounter is therefore a coupled and asymmetric multi-vehicle interaction: one privileged agent operates under altered mobility rules, while multiple civilian agents adapt to its presence. The safety-critical phenomenon is the resulting traffic-wide behavioral change, leading to increasing mobility risks.

These induced behavioral changes create a broader evaluation problem for autonomous-driving and transportation safety research. Models and control strategies are commonly developed under nominal assumptions about lane use, vehicle interaction, and right-of-way, whereas an EMV encounter can alter all three simultaneously. Its effects may therefore extend across perception, motion forecasting, interaction modeling, and safety assessment. Determining how such systems respond requires more than collecting isolated appearances of emergency vehicles. It requires reproducible interaction data in which EMV privileges, civilian responses, and traffic conditions can be configured while the simulated environment, sensing, and ground truth remain consistent.

Existing resources provide only parts of this capability. Naturalistic datasets such as nuScenes and the Waymo Open Dataset \cite{caesar2020nuscenes,sun2020wod}, including recent long-tail extensions \cite{xu2025wode2e,huang2026nureasoning}, provide realistic observations but only for encounters that happened to be recorded; they offer no direct control over EMV behavior or civilian response. Transportation simulators such as SUMO and VISSIM support studies of EMV routing, signal priority, and response time \cite{lopez2018sumo,su2022emvlight}, but are aimed chiefly at operational analysis rather than multimodal AI evaluation. CARLA-based benchmarks provide configurable sensing and interactive simulation \cite{dosovitskiy2017carla,jia2024bench2drive}, yet are typically organized around a finite suite of scenario definitions with actor behavior specified inside each scenario. What remains missing is behavior-level control of EMV--civilian interactions, through which EMV privileges and civilian response policies can be configured while synchronized sensor observations and simulator-native ground truth are collected for downstream evaluation.

We address this gap with \textbf{SIREN}, a behavior-driven SUMO--CARLA co-simulation platform for generating EMV interactions. SIREN couples SUMO's network-level traffic evolution and behavior logic with CARLA's continuous vehicle control and synchronized onboard sensing. Depending on the active behavior, the EMV--civilian interaction is controlled by SUMO, CARLA, or jointly through synchronized state exchange and control transfer. Neither simulator alone provides this complete behavior-to-data loop: SUMO enables coordinated traffic responses, while CARLA realizes them as continuous motion and produces the sensor-rich observations needed for benchmark construction. Together, they generate behavior-conditioned trajectories with simulator-native actor states, 3D geometry, traffic-control states, and interaction events.

\begin{figure}[t]
    \centering
    \includegraphics[width=\textwidth]{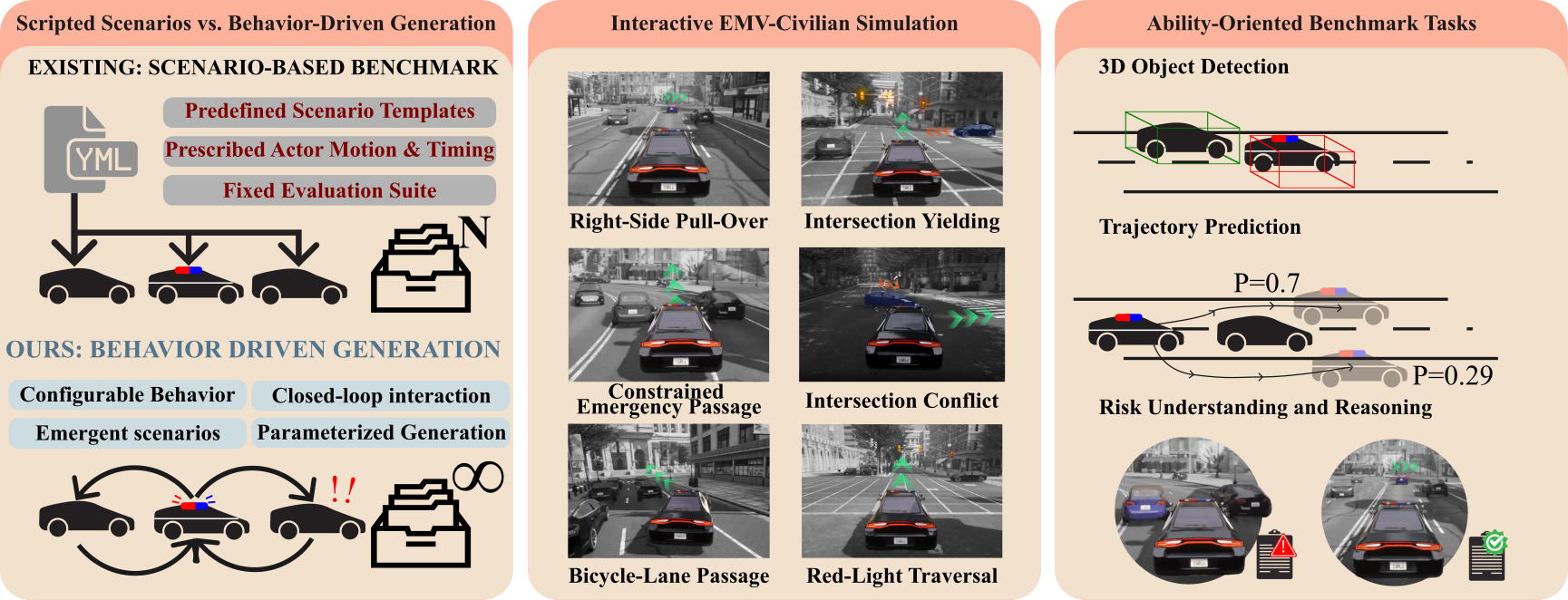}
    \caption{
    Overview of SIREN and SIREN-Bench.
    (1) Existing scenario-based benchmarks typically evaluate a fixed set of predefined scenarios with prescribed actor motion and timing. SIREN instead specifies EMV and civilian behavior policies and generates scenario realizations through closed-loop interaction.
    (2) Representative EMV--civilian interactions generated by the SIREN SUMO--CARLA co-simulation. Reading in row order, the panels illustrate the rescue-corridor pull-over and lane-clearing responses of the clearance family (S1--S2), privileged traversal of controlled intersections (S3--S6), and use of nonstandard road space (S7); template identifiers are defined in Table~\ref{tab:scenarios}.
    (3) SIREN-Bench evaluates three complementary AI capabilities: 3D object detection, trajectory prediction under EMV-induced interaction, and scene-level risk understanding from onboard visual observations.
    }
    \label{fig:concept}
\end{figure}

To facilitate behavior-driven data generation and evaluation, we instantiate SIREN as \textbf{SIREN-Bench-v1}. We configure seven parameterized EMV interaction templates spanning emergency levels L1--L3, from privileged EMV motion without induced civilian yielding to lane-clearing and rescue-corridor responses. The templates cover three behavior families: traffic clearance, privileged traversal of traffic controls, and nonstandard road-space use. The platform supports variation within each template. For 3D object detection and trajectory prediction, v1 uses one recorded base episode per template, so all methods within those tasks operate on the same seven episodes. Risk understanding is evaluated separately on 105 front-camera videos, with 15 videos per template. As representative uses of this benchmark, we select three downstream tasks (Figure~\ref{fig:concept}): \textbf{3D object detection}, \textbf{trajectory prediction}, and \textbf{vision-language risk understanding}, which respectively probe perception, motion forecasting, and scene-level risk inference under EMV-induced behavior. SIREN can also be extended to other tasks, such as behavior modeling, planning and control, traffic-flow analysis, and safety assessment.

Experiments with nine trajectory predictors, four LiDAR-based 3D detectors, and five vision-language models reveal different failure patterns across tasks. Detection is weakest in traffic-clearance episodes, where yielding creates dense and off-center vehicle configurations. Prediction is weakest during privileged intersection traversal, especially at stop-controlled intersections, and no learned predictor outperforms the constant-velocity reference on average. Vision-language models exhibit strong class-dependent biases: no evaluated model obtains non-zero F1 for all three risk classes, and the highest-accuracy model fails to identify any Near-Miss or Collision cases. Thus, unconventional EMV motion alone does not determine difficulty; the behavior it induces in surrounding traffic matters, and it affects perception, prediction, and semantic understanding differently.

Our contributions are threefold:
\begin{itemize}\itemsep2pt \parskip0pt
    \item \textbf{Behavior-centered benchmarking.} We introduce a benchmarking perspective in which EMV privileges and induced civilian responses are the primary factors for data generation and evaluation, rather than treating each scenario as an isolated test case.

    \item \textbf{Hybrid SUMO--CARLA co-simulation.} We develop SIREN, which uses synchronized SUMO--CARLA control to generate behavior-level EMV--civilian interactions together with onboard sensor observations and simulator-native annotations.

    \item \textbf{SIREN-Bench-v1 and representative evaluations.} We configure seven interaction templates spanning emergency levels L1--L3 and evaluate 3D object detection, trajectory prediction, and vision-language risk understanding, showing that different behaviors challenge different tasks.
\end{itemize}

The remainder of the paper reviews related work, presents the SIREN behavior model and co-simulation pipeline, describes the construction of SIREN-Bench-v1, and evaluates the three downstream tasks before concluding with the limitations of the current release.

\section{Related Work}

\subsection{Emergency-Vehicle Traffic Simulation and Control}
SUMO provides the microscopic substrate for modeling vehicles, signals, demand, and simulator coupling \cite{lopez2018sumo}, and within this tradition Bieker-Walz et al. model EMV special rights including blue-light activation, traffic-rule exceptions, and surrounding traffic that forms an emergency lane \cite{bieker-walz_analysis_nodate}. Most such work treats emergency mobility as a transportation operations problem: reducing response time, improving routing, or adjusting signals so an EMV can move through congested traffic. The closest precedent for EMV-civilian interaction is Cortes and Stefoni, who model civilian lane changes, sidewalk mounting, intersection approach, and traffic-light handling through a Paramics API calibrated on Santiago fire-truck video and GPS \cite{cortes_trajectory_2023}. Other work targets optimization or control, such as EMVLight for decentralized multi-agent RL over coupled EMV routing and signal preemption \cite{su_emvlight_2023}, tactical decision-making via rule-based avoidance plus a speed-adaptive DQN \cite{niu_tactical_2021}, and adaptive motion control via online policy meta-learning \cite{lei23cola}. These works show EMV behavior and civilian response can be modeled, but their objective remains traffic operation rather than benchmark generation for autonomous driving.

\subsection{Autonomous-Driving Benchmarks and Simulators}
CARLA provides an open urban-driving simulator for configurable sensors and benchmark tasks \cite{dosovitskiy2017carla}, while OpenCDA is the most direct co-simulation precedent, integrating CARLA and SUMO for cooperative-driving automation with a scenario manager \cite{xu_opencda_2023}. Alongside this separate line of simulators, large datasets and scenario-generation surveys establish the need for scalable evaluation data and rare-event test cases \cite{caesar2020nuscenes,sun2020wod,ding_survey_2023}, and recent closed-loop benchmarks improve realism in different ways: nuPlan on real-world data \cite{caesar_nuplan_2022}, Bench2Drive on end-to-end driving abilities in CARLA \cite{jia2024bench2drive}, Waymax on accelerated data-driven multi-agent simulation \cite{gulino_waymax_2023}, and NAVSIM trading reactivity for scale \cite{dauner_navsim_2024}. The strongest recent attempt at long-tail event data is HiDrive, which explicitly includes emergency-vehicle yielding and red-light emergency-yielding scenarios \cite{xia_hidrive_2026}. Even there, routes and scenario categories are pre-specified within a finite curated suite, with no EMV behavior model to generate emergent interactions. Existing benchmarks thus include EMV response as scenario content, but not as the mechanism for generating new benchmark instances.

\section{The SIREN Platform}
\label{sec:platform}

\begin{figure}[t]
    \centering
    \includegraphics[width=0.92\textwidth]{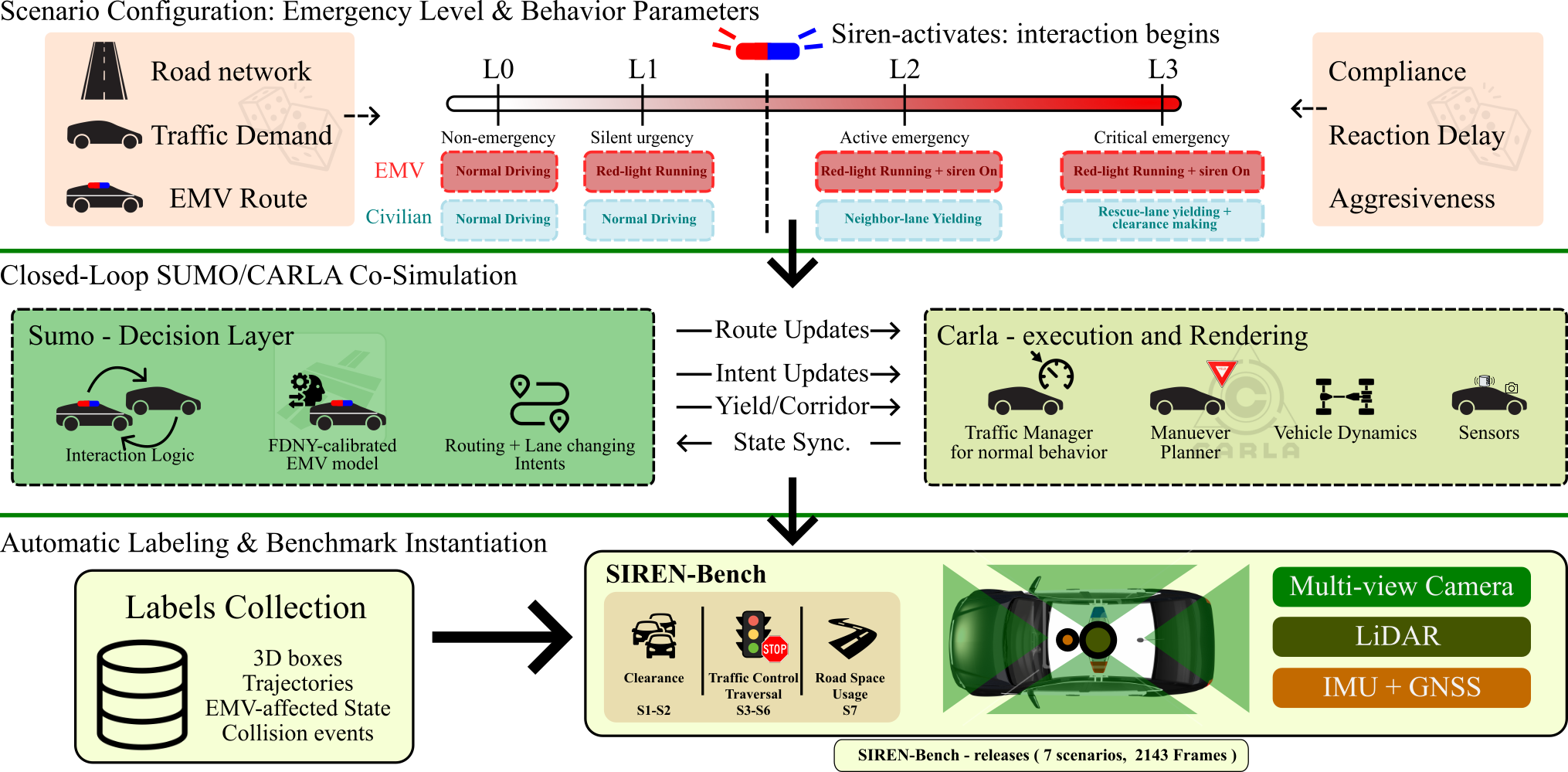}
    \caption{Overview of the SIREN behavior-driven generation pipeline. Rather than prescribing complete scenarios, SIREN configures EMV and civilian behaviors whose closed-loop interaction is resolved through SUMO--CARLA co-simulation. Simulator-native states and sensor observations are then used to construct SIREN-Bench for downstream evaluation. The emergency-level axis (L0--L3) is defined in Table~\ref{tab:levels}; SIREN-Bench-v1 releases the L1--L3 settings as seven templates, with a risk-understanding set of 105 videos (15 per template and 190 frames per video; 19{,}950 frames in total).}
    \label{fig:pipeline}
\end{figure}

\begin{figure}[t]
    \centering
    \includegraphics[width=0.85\textwidth]{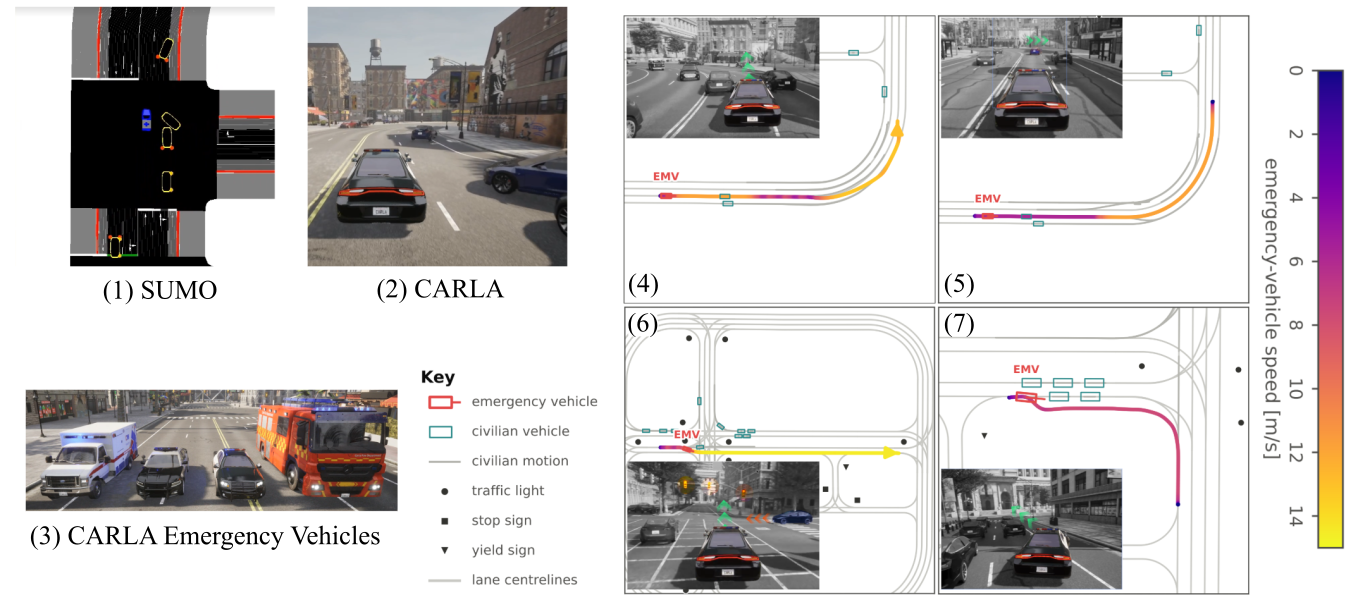}
    \caption{Illustrations of SIREN.
(1) SUMO view of the synchronized traffic state.
(2) CARLA rendering of the corresponding scene.
(3) Supported EMV assets.
(4) EMV passes between queued vehicles (template S1).
(5) Civilian lane clearing for EMV passage (S2).
(6) Privileged EMV traversal at a controlled intersection (S3--S6).
(7) EMV use of nonstandard road space while turning (S7).
Trajectory color indicates EMV speed.}
    \label{fig:platform_examples}
\end{figure}

Configuration in SIREN begins at the behavior layer, which sets the privileged EMV policy and the civilian response for a given emergency level and traffic condition. A closed-loop SUMO--CARLA co-simulation then resolves these behaviors into multi-agent traffic interactions while generating synchronized onboard observations, and the resulting simulator states are converted into task-specific annotations and benchmark episodes. Figure~\ref{fig:pipeline} summarizes the three stages. Because behaviors are specified before trajectories, individual scenarios emerge from the interaction between the EMV policy, civilian responses, and the surrounding traffic state. Figure~\ref{fig:platform_examples} shows representative views of the platform, the EMV assets it supports, and four of the interactions it generates, illustrating how privileged EMV behavior combines with traffic-dependent civilian responses to produce diverse outcomes rather than replaying fixed trajectories.

\subsection{Behavior-Driven Interaction Model}
\label{sec:behavior}

The calibrated microscopic EMV behavior model of Zuo et al. \cite{zuo_hybrid_2026}, obtained with a hybrid evolutionary-algorithm and reinforcement-learning (EA--RL) procedure, supplies the longitudinal and gap-acceptance parameters of a privileged EMV inside SUMO. SIREN builds on it. That model does not specify how civilians respond, however, and it has no sensing or rendering component. Our additions are the level-dependent civilian response policies described below, the event-based SUMO--CARLA control-transfer scheme of the next section, and the conversion of the resulting rollouts into annotated benchmark episodes.

Four emergency levels (Table~\ref{tab:levels}) form the configuration axis of Figure~\ref{fig:pipeline}. L0 is a non-emergency control setting in which the EMV holds no privilege and civilians follow nominal traffic rules; it exists so that the same road network, demand, and route can be replayed without an emergency, and is not instantiated in SIREN-Bench-v1. Levels L1--L3 share the same privileged EMV policy, including traffic-control traversal and SUMO-derived lateral maneuvers, and differ in the emergency cue and the response induced in surrounding traffic. At L1, the siren is inactive and civilians remain under nominal traffic control, allowing privileged EMV motion without induced yielding. L2 activates \emph{lane-clearing yield}, in which affected vehicles perform a feasible lane change to vacate the EMV path. L3 activates \emph{rescue-corridor yield}, where vehicles move toward their lane boundaries and hold the resulting clearance position until the EMV passes.

\begin{table}[t]
\caption{Level-Dependent Control in SIREN}
\centering
\sirenTableStyle
\newcolumntype{R}[1]{>{\raggedright\arraybackslash}p{#1}}
\begin{tabular}{@{}l c R{2.6cm} R{3.2cm} l@{}}
\toprule
Level & Siren & \makecell[l]{EMV route\\mode} & \makecell[l]{Civilian\\response} & \makecell[l]{Tem-\\plates} \\
\midrule
L0 & off & Nominal; no privilege & Nominal traffic control & --- \\
L1 & off & Next-link; intent override & Nominal traffic control & S3--S6 \\
L2 & on  & As L1 & Feasible yield lane change & S2 \\
L3 & on  & Next-link/intent; copied SUMO route; gap escape & Pull over, then braking hold & S1, S7 \\
\bottomrule
\end{tabular}

\vspace{4pt}
{\footnotesize Note: L0 is a non-emergency control setting and is not instantiated in
SIREN-Bench-v1. Levels L1--L3 share the same EMV privilege policy (stop- and
red-signal traversal and nonstandard road-space use on shoulder lanes), so they
vary only the emergency cue and the induced civilian response. An intent override
is a temporary CARLA lane-change path derived from SUMO's lateral request.}
\label{tab:levels}
\end{table}

Once activated, an assigned response persists as an actor state until the EMV has passed; it is not a one-step command. A requested maneuver that is temporarily infeasible is deferred and reconsidered at subsequent simulation steps. Response delay, incomplete lane clearing, and corridor formation can therefore arise from the evolving traffic state instead of from prescribed actor trajectories.

\subsection{Hybrid SUMO--CARLA Co-Simulation}
\label{sec:cosim}

SIREN uses a moving, fixed-radius interaction region centered on the EMV. Traffic outside this circle evolves primarily in SUMO; actors entering it join the CARLA-focused synchronization and control set. At every synchronous step SUMO supplies the EMV's next-link sequence, left/right/sublane lane-change states, and per-vehicle \texttt{rescueLane} status, plus, at L3, rescue-corridor waypoints and the leader information used to detect a blocked EMV. CARLA converts the selected path into continuous throttle, brake, and steering through a shared waypoint navigator whose waypoint source, target speed, and collision handling change with the active mode. All lateral states are monitored, but only valid requests are executed; infeasible ones are deferred and reconsidered.

Under L1--L2 the EMV follows its next-link route, taking a temporary local path when SUMO requests a feasible lane change, and L1 civilians remain under CARLA Traffic Manager. At L2 a yield-active civilian is handed to a collision-aware lane-change navigator, retaining its original path so nominal control can resume after release. At L3 civilians use a pull-over navigator and then hold position, while the EMV follows a copied SUMO corridor path; this mode suppresses new intent-driven lane changes to prevent conflicting route sources. Should leader and speed evidence show the EMV still blocked, a higher-priority \emph{gap-escape} planner builds a local recovery path and temporarily limits collision avoidance to permit controlled passage. Control ownership thus shifts with the level. The code-level arbitration, waypoint controller, and gap-recovery search are detailed in Section~\ref{sup:behavior}. The event-based scheme keeps global traffic consistency in SUMO while letting CARLA realize short-horizon, interaction-specific motion without prescribing complete trajectories. Each synchronized rollout records actor poses, kinematics, controller states, signal phases, and calibrated onboard observations for all SIREN-Bench tasks.

\subsection{Benchmark Construction}
\label{sec:construction}

Each of the seven templates in SIREN-Bench-v1 defines a target interaction and road configuration, while the multi-agent trajectories are resolved through the closed-loop behavior model instead of being prescribed. The templates are parameterized and span three behavior families: traffic clearance, traffic-control traversal, and nonstandard road-space use (Table~\ref{tab:scenarios}). Scenario-specific parameters (EMV and background speeds, traffic volume, response distance, maneuver timing) can be varied to produce different realizations of the same interaction. In v1, detection and trajectory prediction use one recorded base episode per template, whereas risk understanding uses a separate set of 15 front-camera videos per template. All evaluated methods within a task receive the same task-specific data.

All base episodes are generated on the CARLA \texttt{Town10HD\_Opt} map, last 25--45~s, and are logged at 10~Hz, giving 250--450 actor-state frames per episode. The risk-understanding set comprises 105 front-camera videos---15 for each of the seven templates---with 190 frames per video and 19{,}950 video frames in total. Sensing uses a fixed EMV-mounted suite: a 64-beam LiDAR at 1.93~m with 120~m range, $1.3\times10^{6}$ points/s, 20~Hz rotation and a $+5^{\circ}/-25^{\circ}$ vertical field of view; four $800\times600$ RGB cameras at $100^{\circ}$ field of view (front, rear, left/right rear-facing); and co-located GNSS and IMU. This LiDAR is denser than the 32-beam sensor used in nuScenes, one reason the pretrained detectors below should not be compared against their published nuScenes scores.

\begin{table}[t]
\caption{Seven Parameterized Interaction Templates in SIREN-Bench-v1}
\centering
\sirenTableStyle
\begin{tabular}{@{}c l p{7.2cm} c@{}}
\toprule
ID & Behavior family & Instantiated interaction & $N_{\mathrm{veh}}$ \\
\midrule
S1 & Clearance &
EMV passes between two queues of vehicles. & 30 \\

S2 & Clearance &
Vehicles merge out of the EMV lane to create a clear passage. & 20 \\

S3 & Traffic control &
EMV traverses a red signal with same-direction traffic in the intersection. & 20 \\

S4 & Traffic control &
EMV traverses a red signal with crossing traffic in the intersection. & 20 \\

S5 & Traffic control &
EMV traverses a stop sign with same-direction traffic in the intersection. & 20 \\

S6 & Traffic control &
EMV traverses a stop sign with crossing traffic in the intersection. & 30 \\

S7 & Road-space use &
EMV occupies the lane shoulder while executing a right turn. & 20 \\
\bottomrule
\end{tabular}

\vspace{4pt}
{\footnotesize Note: Templates are organized by the EMV behavior they instantiate.
$N_{\mathrm{veh}}$ is the configured background-traffic count.}
\label{tab:scenarios}
\end{table}

Object-level annotations for the downstream tasks come directly from the synchronized sensor observations and simulator states recorded for each episode, which include actor trajectories, kinematics, traffic-signal states, and camera calibration. Because CARLA exposes no vehicle sub-category for the detection labels, every vehicle, including the EMV, is exported as a single \texttt{Car} class; the benchmark therefore measures how the EMV interaction perturbs vehicle detection as a whole, and does not yet report a separate EMV-class AP. The seven base episodes support the trajectory-prediction and perception evaluations, while risk understanding uses the separate 105-video set described above.

\section{Experiments}

\subsection{Experimental Setup}

We generate the seven interaction templates of Table~\ref{tab:scenarios} on a compute cluster, allocating each simulation job four CPU cores, 32~GB of memory, and one 32-GB NVIDIA Tesla V100S GPU. CARLA~0.9.16 runs in synchronous mode with a fixed simulation step of 0.05~s (20~Hz) and renders camera observations at $800\times600$ pixels, while SUMO maintains the traffic and interaction states described in the co-simulation section above; actor states are logged at 10~Hz. Unless otherwise stated, all learned models use their released pretrained checkpoints without fine-tuning on SIREN-Bench. Further implementation details appear in Sections~\ref{sup:cosim}--\ref{sup:proto}.

\subsubsection{Trajectory prediction}
Seven learned predictors are compared: CSP \cite{deo2018csp}, STDAN \cite{chen2022stdan}, BAT \cite{liao2024bat}, EMP-M and EMP-D \cite{prutsch2024emp}, DeMo \cite{zhang_demo_2024}, and DeMo with RealMotion \cite{song2024realmotion}; constant velocity (CV) and the Intelligent Driver Model (IDM) are non-learned references.

All methods are scored on a common grid: a 5-s future at 0.2~s (25 steps), one sample per (vehicle, frame) pair with a fully observed window, stride one frame. \emph{Every} qualifying vehicle is scored, including the EMV; v1 reports no per-agent-category breakdown. Observation windows differ by family, since each released checkpoint fixes them: CSP, STDAN, BAT and the non-learned references take a 3-s history at 0.2~s (NGSIM convention), whereas EMP-M, EMP-D and DeMo take a 5-s history at 0.1~s and natively emit a 6-s future, which we resample onto the common grid. History length is thus a confound between families, and within-family comparisons are the more reliable ones. Stride-one sampling leaves the samples strongly overlapping. The templates yield 7{,}062 scored samples (914/1{,}371/825/1{,}247/605/340/1{,}760 for S1--S7) drawn from far fewer distinct behaviors, so they are not independent for significance testing. We report ADE and FDE here; minADE$_6$, minFDE$_6$, and 5-s RMSE are in the supplementary material.

\subsubsection{3D object detection}
We evaluate four representative LiDAR-based detectors pretrained on nuScenes: PointPillars \cite{lang2019pointpillars}, SECOND \cite{yan2018second}, VoxelNeXt \cite{chen2023voxelnext}, and TransFusion-L \cite{bai2022transfusion}. Scoring follows the official nuScenes protocol \cite{caesar2020nuscenes}: matches are resolved by bird's-eye-view center distance rather than 3D IoU, mAP averages AP over the $\{0.5,1,2,4\}$~m thresholds, and NDS combines mAP with the true-positive error terms. All vehicles carry a single \texttt{Car} label, as noted above, so mAP here measures vehicle detection in the presence of the EMV interaction rather than EMV recognition specifically. The detailed true-positive error breakdown is deferred to the supplement.

\subsubsection{Risk understanding}
Inspired by \textit{SeeUnsafe}~\cite{zhang2025seeunsafe}, we use its
preprocessing and prompting components unchanged, including segmentation and
visual grounding as visual prompts together with the accompanying text prompt.
We apply this pipeline to the SIREN-Bench videos and reference labels.
Vision-language risk understanding evaluates whether a VLM can infer the
scene-level risk category from an EMV--civilian interaction. Although this
inference requires interpreting the agents and their interactions, the present
benchmark evaluates the resulting classification rather than the model's
intermediate reasoning process.
We evaluate five
vision-language models: Blaifa-InternVL3.5-8B~\cite{wang2025internvl35},
Gemma3-12B~\cite{gemma_2025},
LLaVA-Llama3-8B~\cite{2023xtuner},
MiniCPM-V-4.5-8B~\cite{yu_minicpm-v_2025}, and
Qwen3.5-9B~\cite{qwen2026qwen35}. Each of the 105 MP4 videos produces two
intermediate predictions over the classes Normal, Near-Miss, and Collision.
Following SeeUnsafe, the more severe of the two predictions is selected with
a max-severity rule (Normal $<$ Near-Miss $<$ Collision), yielding one final
video-level prediction per video.
Reference labels are derived from the recorded interactions and collision
events. We report overall classification accuracy together with per-class
precision and F1.

\begin{table}[t]
\centering
\caption{Trajectory Prediction on SIREN-Bench}
\label{tab:prediction}
\resizebox{\textwidth}{!}{%
\begin{tabular}{lcccccccc}
\toprule
& \multicolumn{2}{c}{Clearance}
& \multicolumn{4}{c}{Traffic-control traversal}
& \multicolumn{1}{c}{Road-space} & \\
\cmidrule(lr){2-3}\cmidrule(lr){4-7}\cmidrule(lr){8-8}
Model & S1 & S2 & S3 & S4 & S5 & S6 & S7 & Mean \\
\midrule
CV
& 1.63/4.30 & 1.17/3.11 & 0.96/2.40 & 0.68/1.65
& 1.95/5.03 & 2.86/7.56 & 0.85/2.08 & 1.44/3.73 \\

IDM
& 2.23/5.14 & 1.58/3.75 & 1.18/2.69 & 1.04/2.18
& 2.68/6.49 & 3.29/8.34 & 1.06/2.55 & 1.86/4.45 \\
\midrule

CSP
& 2.63/6.12 & 3.57/7.79 & 2.92/6.65 & 2.54/5.83
& 3.25/7.10 & 4.63/9.69 & 1.82/4.02 & 3.05/6.74 \\

STDAN
& 3.99/8.12 & 6.02/12.13 & 4.20/9.14 & 3.11/6.81
& 4.58/9.94 & 7.87/15.41 & 2.38/5.93 & 4.59/9.64 \\

BAT
& 9.14/17.35 & 9.55/16.14 & 8.17/15.75 & 6.06/10.52
& 4.80/7.85 & 8.40/13.16 & 3.42/5.06 & 7.08/12.26 \\

EMP-M
& 1.79/4.73 & 1.55/4.38 & 2.71/8.24 & 1.88/5.92
& 3.47/9.44 & 3.24/8.86 & 2.05/6.50 & 2.38/6.87 \\

EMP-D
& 1.67/4.44 & 1.32/3.57 & 2.41/7.39 & 1.39/4.60
& 2.13/6.21 & 3.50/9.68 & 1.63/5.83 & 2.01/5.96 \\

DeMo
& 1.65/4.47 & 1.58/4.60 & 2.26/7.00 & 1.26/4.08
& 2.72/7.80 & 3.57/9.96 & 1.05/4.03 & 2.01/5.99 \\

DeMo+RealMotion
& 1.62/4.29 & 1.43/4.22 & 2.35/6.95 & 1.64/4.83
& 2.41/7.03 & 2.50/6.85 & 1.23/3.77 & 1.88/5.42 \\
\bottomrule
\end{tabular}}

\vspace{4pt}
{\footnotesize Note: Each entry reports ADE/FDE (m) over a 5-s prediction horizon;
lower is better.}
\end{table}

\begin{table}[t]
\centering
\caption{3D Object Detection on SIREN-Bench}
\label{tab:detection}
\resizebox{\textwidth}{!}{%
\begin{tabular}{lcccccccc}
\toprule
& \multicolumn{2}{c}{Clearance}
& \multicolumn{4}{c}{Traffic-control traversal}
& \multicolumn{1}{c}{Road-space} & \\
\cmidrule(lr){2-3}\cmidrule(lr){4-7}\cmidrule(lr){8-8}
Model & S1 & S2 & S3 & S4 & S5 & S6 & S7 & Mean \\
\midrule
PointPillars
& .398/.407 & .434/.436 & .583/.525 & .589/.549
& .451/.456 & .530/.540 & .569/.528 & .508/.492 \\

SECOND
& .420/.420 & .388/.438 & .631/.563 & .633/.574
& .461/.480 & .523/.540 & .590/.556 & .521/.510 \\

VoxelNeXt
& .438/.393 & .413/.434 & .641/.571 & .616/.558
& .468/.454 & .568/.540 & .578/.543 & .532/.499 \\

TransFusion-L
& .424/.411 & .423/.456 & .648/.566 & .618/.565
& .491/.481 & .562/.555 & .610/.571 & .539/.515 \\
\bottomrule
\end{tabular}}

\vspace{4pt}
{\footnotesize Note: Each entry reports mAP/NDS for the Car class using
nuScenes-pretrained checkpoints; higher is better.}
\end{table}

\subsection{Main Benchmark Evaluation}\label{sec:main-eval}

\subsubsection{Trajectory prediction}
Averaged over the seven templates, the best learned predictor in Table~\ref{tab:prediction} reaches 1.88~m ADE, against 1.44~m for constant velocity and 1.86~m for IDM. Although a learned model slightly outperforms the constant-velocity reference on S1 and S6 individually, neither non-learned reference is surpassed on the seven-template mean. The released checkpoints therefore do not transfer favorably to SIREN-Bench under the present evaluation setup. This comparison does not isolate a single cause: the gaps may reflect the EMV-induced interactions, source-domain shift, fixed differences in observation history, or model-specific input and adaptation choices. Among the learned methods, the Argoverse-trained models (EMP and DeMo) achieve lower mean ADE than the NGSIM-trained models (CSP, STDAN, and BAT).

The clearance scenarios S1--S2 introduce coordinated lateral responses as surrounding vehicles create space for the EMV, and they produce large errors for several predictors. More recent methods stay closer to the non-learned references, indicating that sustained lane-clearing motion can be captured when sufficient interaction context is available. Prediction performance, in short, depends strongly on the type of EMV interaction.

S6, where the EMV traverses a stop-controlled intersection with crossing traffic, is the highest-error scenario for seven of the nine evaluated methods. Averaged over the learned predictors, ADE reaches 4.82~m in S6, compared with 1.94~m in S7. For the stronger predictors, S5--S6 (stop-controlled) also tend to be more difficult than the corresponding red-light cases S3--S4. Traffic-control traversal thus poses a challenge of its own: surrounding vehicles continue to negotiate the intersection while responding to the EMV's privileged motion, producing future trajectories that depart from ordinary right-of-way interactions. S7, by contrast, changes the road space occupied by the EMV but induces less coordinated response and is comparatively easy to predict. The primary difficulty for prediction therefore comes from the \emph{interaction induced by the EMV}, not from unconventional EMV motion alone. Two caveats apply. S6 contributes the fewest scored samples (340), so its position as hardest rests on less evidence than the other columns. The learned predictors also come from different source domains (CSP, STDAN and BAT from NGSIM highway car-following, EMP and DeMo from Argoverse~2 urban driving), so the gap between families reflects a training-domain shift as well as the EMV interaction. Our claims rest on the ranking across templates, which is unaffected, since every model sees the same seven.

\subsubsection{3D object detection}
Mean mAP across the four detectors is about 0.42 in S1 and 0.41 in S2, against 0.63 and 0.61 in the red-light templates S3--S4. The observed pattern in Table~\ref{tab:detection} differs from the prediction results: detection scores are consistently lowest in the clearance templates. S1 sends the EMV through the corridor between two queues, and S2 moves several vehicles laterally while clearing a lane; both contain dense, off-center configurations that depart from lane-organized traffic and increase geometric overlap among nearby actors. The lower scores are consistent with these geometric conditions being challenging for zero-shot detection.

The traffic-control templates exhibit higher mAP. S3--S4 retain conventional vehicle placement and give the highest values in the benchmark, and S7 provides a similar contrast: although the EMV leaves the travel lane for the shoulder, mean mAP stays at 0.59, well above S1--S2. The stop-sign templates S5--S6 fall between these extremes at 0.45--0.57; their queued approaches may contribute some of the dense geometry also present in S1--S2. Ordering templates by mean mAP gives clearance (0.42) $<$ stop-sign traversal (0.51) $<$ shoulder use (0.59) $<$ red-light traversal (0.62). In this release, lower mAP therefore coincides with episodes in which surrounding traffic is more strongly reorganized. Because detection uses one base episode per template, this comparison cannot separate the effect of behavior from episode-specific geometry, density, or vehicle placement; randomized realizations are needed to isolate those factors. NDS follows the same observed trend, and the true-positive breakdown is in the supplementary material. Absolute values should not be read against published nuScenes numbers: the checkpoints run zero-shot on a denser 64-beam sensor and a different city layout, and we make no claim about the size of that shift.

\begin{table}[t]
\centering
\caption{Per-Class Risk Classification Results on SIREN-Bench}
\label{tab:risk_classification}
\small
\setlength{\tabcolsep}{4pt}
\renewcommand{\arraystretch}{1.12}

\begin{adjustbox}{max width=\linewidth}
\begin{tabular}{@{}lccccccc@{}}
\toprule
& & \multicolumn{2}{c}{Normal}
& \multicolumn{2}{c}{Near-Miss}
& \multicolumn{2}{c}{Collision} \\
\cmidrule(lr){3-4}
\cmidrule(lr){5-6}
\cmidrule(lr){7-8}
Model & Acc. & Prec. & F1 & Prec. & F1 & Prec. & F1 \\
\midrule
Blaifa-InternVL3.5-8B
& 47.62 & 84.21 & 57.14 & 28.12 & 40.45 & 0.00 & 0.00 \\

Gemma3-12B
& 26.67 & 90.00 & 21.43 & 23.75 & 36.19 & 0.00 & 0.00 \\

LLaVA-Llama3-8B
& 22.86 & 0.00 & 0.00 & 23.08 & 37.21 & 0.00 & 0.00 \\

MiniCPM-V-4.5-8B
& 49.52 & 85.96 & 74.81 & 0.00 & 0.00 & 6.25 & 11.11 \\

Qwen3.5-9B
& \textbf{70.48} & 70.48 & \textbf{82.68}
& 0.00 & 0.00 & 0.00 & 0.00 \\
\bottomrule
\end{tabular}
\end{adjustbox}

\vspace{4pt}
{\footnotesize
Accuracy (Acc.) is reported over the 105 video-level predictions. Precision
(Prec.) and F1 are reported separately for the Normal, Near-Miss, and Collision
classes. Values are percentages; higher is better.
}
\end{table}

\vspace{4pt}

\subsubsection{Risk understanding}
Table~\ref{tab:risk_classification} shows that overall accuracy does not
reliably measure three-class risk identification. Qwen3.5-9B obtains the
highest accuracy (70.48\%) and Normal-class F1 (82.68\%), but has zero
precision and F1 for both Near-Miss and Collision. Its apparent advantage
therefore comes from favoring the dominant Normal class rather than from
discriminating among the three risk levels. LLaVA-Llama3-8B exhibits the
opposite pattern: its 22.86\% accuracy and 37.21\% Near-Miss F1 are paired
with zero Normal and Collision scores, indicating a strong bias toward the
Near-Miss class.

Performance on the safety-critical classes remains weak across all five
models. Blaifa-InternVL3.5-8B attains the highest Near-Miss F1 (40.45\%) but
has zero Collision F1. MiniCPM-V-4.5-8B is the only model with non-zero
Collision performance, yet its Collision precision and F1 are only 6.25\%
and 11.11\%, and its Near-Miss F1 is zero. Gemma3-12B achieves 90.00\%
Normal precision but only 21.43\% Normal F1 and no correct Collision
identification. No evaluated model obtains non-zero F1 for all three
classes.

Taken together, the models vary not only in overall performance but also in
their characteristic error mode. Some default toward Normal, whereas others
are overly sensitive to risky driving behavior and favor Near-Miss; none
balances all three classes reliably. SIREN-Bench makes these opposing
tendencies visible by deliberately presenting EMV--civilian interactions
that can develop into Normal, Near-Miss, or Collision outcomes. Such
privileged-agent interactions are uncommon in conventional driving data, but
the repeated videos across SIREN-Bench's seven templates provide a focused
test of how VLMs interpret them. The aggregate results expose model-level
biases; per-template confusion matrices would additionally be needed to
attribute those biases to particular EMV behaviors.

\section{Conclusion}
\label{sec:conclusion}

We presented SIREN, a behavior-driven platform for generating emergency-vehicle interactions through closed-loop SUMO--CARLA co-simulation, and instantiated it as SIREN-Bench-v1 with seven parameterized interaction templates. Rather than prescribing complete scenarios, SIREN specifies privileged EMV behavior and adaptive civilian responses and lets the resulting interactions unfold with the surrounding traffic. Evaluations on trajectory prediction, 3D object detection, and risk understanding show that different EMV behaviors stress different aspects of existing models. Traffic-clearance episodes show the lowest detection scores and coincide with strongly reorganized nearby traffic, while privileged intersection traversal creates greater difficulty for motion prediction; risk-understanding models exhibit strong class-dependent biases, with no evaluated model obtaining non-zero F1 across all three risk categories. These results motivate behavior-driven generation as a complementary approach to fixed scenario suites for studying how transportation algorithms respond to privileged-agent interactions.

SIREN-Bench-v1 represents an initial instantiation of the platform. Each interaction template is currently evaluated at a single emergency level and on a single map. Detection and trajectory prediction use one base realization per template, which limits analysis across repeated seeds, while risk understanding is reported only at the aggregate benchmark level. These choices prevent separating level, map, and realization effects. Detection evaluation also uses the generic \texttt{Car} class rather than EMV-specific labels, and the current release uses only onboard sensing. Future work will evaluate the same interaction templates across behavior levels, maps, and randomized realizations, introduce EMV-specific and roadside annotations, and quantify the diversity induced by the platform's behavioral parameters.

\clearpage
\appendix
\section{Appendix}
\noindent
This supplement is organized in two parts, mirroring the two artifacts
introduced in the main paper: the SIREN platform and the SIREN-Bench benchmark. Part~A documents the \textbf{SIREN platform}: the
co-simulation loop and control ownership (Sec.~\ref{sup:cosim}), the onboard
sensor suite and placement (Sec.~\ref{sup:sensors}), the behavior
implementation as per-step control logic (Sec.~\ref{sup:behavior}), the
per-mode controller target speeds (Sec.~\ref{sup:speeds}), and the data formats
and coordinate conventions (Sec.~\ref{sup:formats}). Part~B documents the
\textbf{SIREN-Bench evaluation}: the metric definitions (Sec.~\ref{sup:metrics}),
the task-level protocols and model setup (Sec.~\ref{sup:proto}), model and
checkpoint provenance (Sec.~\ref{sup:provenance}), extended
trajectory-prediction metrics (Sec.~\ref{sup:traj}), and the detection
true-positive error breakdown (Sec.~\ref{sup:det}). It is self-contained and
does not introduce any additional results except those summarized in the main paper.

% ##########################################################################
% PART A --- THE SIREN PLATFORM
% ##########################################################################
\subsection{The SIREN Platform}
\label{sup:platform}

\subsubsection{Co-Simulation Loop and Control Ownership}
\label{sup:cosim}

SIREN advances SUMO and CARLA in lockstep. All base episodes are generated on the
CARLA \texttt{Town10HD\_Opt} map with CARLA~0.9.16 and SUMO~1.24.0 in
synchronous mode at a fixed 0.05~s step (20~Hz); actor states are logged at
10~Hz, giving 250--450 actor-state frames per 25--45~s episode. Detection and
trajectory prediction use one such base episode per template. The
risk-understanding set is organized separately as 105 front-camera videos,
with 15 videos per template and 190 frames per video, for 19{,}950 video
frames in total. On a
compute cluster, each simulation job is allocated four CPU cores, 32~GB of
memory, and one 32-GB NVIDIA Tesla V100S GPU.

The simulator boundary is a moving, fixed-radius interaction region centered
on the EMV (a 40~m inner / 50~m outer hysteresis band; Table~\ref{tab:thresholds}).
Traffic outside the circle evolves primarily in SUMO; actors
entering the region join the CARLA-focused synchronization and control set. At each
synchronous step, SUMO supplies the EMV's next-link route sequence
(\texttt{getNextLink}), its left/right/sublane lane-change intent
(\texttt{getLaneChangeIntent}), and per-vehicle \texttt{rescueLane} status,
plus, at level~L3, rescue-corridor waypoints and the leader information used to
detect a blocked EMV. CARLA converts the selected path into continuous
throttle, brake, and steering through a shared \texttt{WaypointNavigator}
whose waypoint source, target speed, and collision handling change with the
active mode. All lateral requests are monitored, but only feasible ones are
executed; infeasible requests are deferred and reconsidered at later steps.

Control ownership shifts with the emergency level (Table~\ref{tab:ownership}).
The privileged EMV policy (stop- and red-signal traversal and nonstandard
road-space use) is shared across L1--L3; the levels differ in the emergency
cue and the civilian response they induce. L0 is a non-emergency control
setting and is not released in SIREN-Bench-v1.

\begin{table}[htbp]
\centering
\caption{Control ownership by emergency level. ``TM'' is the CARLA Traffic
Manager; ``WPN'' is the shared \texttt{WaypointNavigator}. Levels L1--L3 share
the same privileged EMV policy and differ in the civilian response.}
\label{tab:ownership}

\small
\setlength{\tabcolsep}{4pt}
\renewcommand{\arraystretch}{1.12}

\begin{adjustbox}{max width=\linewidth}
\begin{tabular}{@{}llll@{}}
\toprule
Level & EMV controller & Civilian controller & Siren \\
\midrule
L0 & TM (no privilege)
   & TM (nominal)
   & off \\

L1 & WPN (route + intent)
   & TM (nominal)
   & off \\

L2 & WPN (route + intent)
   & WPN (feasible-yield LC)
   & on \\

L3 & WPN (\texttt{copy\_sumo}, \texttt{gap\_escape})
   & WPN (pull-over, hold)
   & on \\
\bottomrule
\end{tabular}
\end{adjustbox}
\end{table}

\subsubsection{Onboard Sensor Suite and Placement}
\label{sup:sensors}

Sensing uses a fixed EMV-mounted suite whose extrinsics are given in
Table~\ref{tab:sensors}, expressed in the CARLA vehicle frame ($x$ forward,
$y$ right, $z$ up, meters; yaw in degrees). The LiDAR, GNSS, and IMU are
co-located at $z=1.93$~m, and the four RGB cameras give front, rear, and
left/right rear-facing coverage; the suite has no pure side-facing camera, so
near-lateral civilian yielding and pull-over are captured only obliquely. The 64-beam LiDAR (120~m range,
$1.3\times10^{6}$ points/s, 20~Hz rotation, $+5^{\circ}/-25^{\circ}$ vertical
field of view) is denser than the 32-beam sensor used in nuScenes, which explains why the pretrained detectors are not directly comparable to their published
nuScenes scores. The released benchmark annotations are based on this suite.
Auxiliary control-only sensors used internally by the generator (a forward
obstacle probe and an RSS sensor) are not part of the released observations.

\begin{table}[htbp]
\centering
\small
\setlength{\tabcolsep}{3pt}
\renewcommand{\arraystretch}{1.12}
\caption{EMV-mounted sensor suite and placement (CARLA vehicle frame:
$x$ forward, $y$ right, $z$ up, in meters; yaw in degrees). All cameras are
$800\times600$ at $100^{\circ}$ horizontal FOV.}
\label{tab:sensors}
\begin{tabular}{@{}lccccl@{}}
\toprule
Sensor & $x$ & $y$ & $z$ & yaw & Key attributes \\
\midrule
LiDAR (64-beam) & 0.0 & 0.0 & 1.93 & 0 & 120\,m, 1.3\,Mpps, 20\,Hz \\
Cam.\ front      & 2.5 & 0.0 & 1.0 & 0    & $800{\times}600$, $100^{\circ}$ \\
Cam.\ right-rear & 0.0 & 0.3 & 1.8 & 100  & $800{\times}600$, $100^{\circ}$ \\
Cam.\ left-rear  & 0.0 & $-0.3$ & 1.8 & $-100$ & $800{\times}600$, $100^{\circ}$ \\
Cam.\ rear       & $-2.0$ & 0.0 & 1.5 & 180  & $800{\times}600$, $100^{\circ}$ \\
GNSS             & 0.0 & 0.0 & 1.93 & 0 & co-located w/ LiDAR \\
IMU              & 0.0 & 0.0 & 1.93 & 0 & accel/gyro noise model \\
\bottomrule
\end{tabular}
\end{table}

\subsubsection{Behavior Implementation: Per-Step Control Logic}
\label{sup:behavior}

Behaviors in SIREN are not scripted trajectories. Each actor runs a control
routine evaluated at every synchronous step over latched flags that persist
across steps, so response delay, incomplete lane clearing, and corridor
formation emerge from the evolving traffic rather than from prescribed motion.
Since a tick evaluates guarded overrides from top to bottom rather than occupying a single state, we present the logic in pseudocode. Algorithm~\ref{alg:emv} is the
level-L3 EMV routine; Algorithm~\ref{alg:npc} is the civilian yield routine
(\texttt{npc\_behavior\_lvl}${=}1$ at emergency level~L2,
\texttt{npc\_behavior\_lvl}${=}2$ at L3). The EMV priority order is
$\texttt{gap\_escape} \succ \texttt{copy\_sumo} \succ \text{forced LC} \succ
\text{route-follow}$: \texttt{gap\_escape} is checked first and, on activation,
cancels an in-progress forced lane change, while \texttt{copy\_sumo} suppresses
new lane-change intent. Predicate names match the controller source.
Algorithm~\ref{alg:emv} shows level~L3; the L1 and L2 EMV routines are the same
routine with the \texttt{gap\_escape} and \texttt{copy\_sumo} branches removed,
leaving next-link route-following with optional intent-driven forced lane
changes.

\begin{algorithm}[t]
\caption{EMV per-step control at level~L3. Latched flags
(\texttt{gap\_escape\_active}, \texttt{lane\_change\_forced}, mode) persist
across ticks; guards are evaluated top-to-bottom. Constants: Table~\ref{tab:thresholds}.}
\label{alg:emv}
\small
\begin{algorithmic}[1]
\REQUIRE \texttt{emergency\_lvl}${=}3$; latched \texttt{emv\_state} persists across ticks
\STATE \textbf{(1) Gap-escape (highest priority)}
\IF{$\mathrm{speed(EMV)} \le v_{\mathrm{stop}}$} \STATE keep stop timer \ELSE \STATE reset stop timer \ENDIF
\STATE $b \gets$ SUMO leader within $80$\,m, else navigator collision snapshot
\IF{\texttt{gap\_escape\_active} and $b{=}\varnothing$ for $\ge t_{\mathrm{grace}}$}
  \STATE \texttt{gap\_escape\_active} $\gets$ false
\ENDIF
\IF{$b{\ne}\varnothing$ and $\mathrm{stopped} \ge t_{\mathrm{stop}}$ and not \texttt{gap\_escape\_active}}
  \STATE \texttt{gap\_escape\_active} $\gets$ true
  \IF{\texttt{lane\_change\_forced}} \STATE \textsc{RestoreLaneChange}() \COMMENT{preempt forced LC} \ENDIF
\ENDIF
\IF{\texttt{gap\_escape\_active}} \STATE replan \textsc{BuildGapEscapeWaypoints}(); disable collision avoidance \ENDIF
\STATE \textbf{(2) Rescue-corridor mode}
\STATE $\mathit{corridor} \gets (\mathit{rescueCount}{>}0 \vee \mathit{brakingAhead}) \wedge \mathit{emvLvl}{\ge}1 \wedge \mathit{npcLvl}{\ge}2$
\IF{$\mathit{corridor}$} \STATE $\mathit{intent} \gets \varnothing$ \COMMENT{copy\_sumo suppresses forced LC} \ENDIF
\STATE \textbf{(3) Forced lane-change}
\IF{not $\mathit{corridor}$ and not \texttt{lane\_change\_forced} and $\mathit{intent}\in\{L,R\}$ and $\mathit{cooldownOK}$}
  \STATE $W \gets$ \textsc{BuildLaneChangeWaypoints}($\mathit{intent}$)
  \IF{$W{\ne}\varnothing$} \STATE \texttt{lane\_change\_forced} $\gets$ true; navigator.\textsc{SetWaypoints}($W$) \ENDIF
\ENDIF
\STATE \textbf{(4) Waypoint source and completion}
\IF{$\mathit{corridor}$ and not \texttt{gap\_escape\_active}} \STATE navigator $\gets$ copied SUMO corridor waypoints \ENDIF
\IF{\texttt{lane\_change\_forced} and (enteredTargetLane $\vee$ navigator.\textsc{Done}() $\vee$ reachedTarget)}
  \STATE \textsc{RestoreLaneChange}()
\ENDIF
\STATE \textbf{(5) Target speed (km/h) by priority}
\IF{\texttt{gap\_escape\_active}} \STATE $\mathrm{set\_speed}(20)$
\ELSIF{$\mathit{corridor}$} \STATE $\mathrm{set\_speed}(30)$
\ELSE \STATE $\mathrm{set\_speed}(50)$ \ENDIF
\STATE apply navigator control
\end{algorithmic}
\end{algorithm}

\begin{algorithm}[t]
\caption{Civilian per-step yield control. \texttt{npc\_behavior\_lvl}${=}1$
gives the feasible-yield lane change (emergency~L2); ${=}2$ gives pull-over
then braking hold (emergency~L3). Latched state persists until the EMV passes.}
\label{alg:npc}
\small
\begin{algorithmic}[1]
\STATE \textbf{Determine yielding (latches until EMV passes)}
\IF{$\mathit{rescueLane}{\ne}\varnothing$}
  \STATE $\mathit{yielding}\gets$ true; latch lateral alignment
\ELSIF{braking or active planner or active target}
  \STATE $\mathit{yielding}\gets \neg\,\textsc{EmvFullyPassed}()$
\ELSE
  \STATE $\mathit{yielding}\gets$ false
\ENDIF
\IF{not $\mathit{yielding}$} \STATE remain under Traffic Manager; \textbf{return} \ENDIF
\STATE \textbf{\texttt{npc\_behavior\_lvl}${=}1$: feasible-yield lane change}
\IF{not \texttt{lane\_change\_forced} and $\mathit{intent}\in\{L,R\}$ and $\mathit{cooldownOK}$}
  \STATE $W \gets$ \textsc{BuildLaneChangeWaypoints}($\mathit{intent}$)
  \IF{$W{\ne}\varnothing$}
    \STATE disable autopilot; store pre-maneuver path; TM speed $-50\%$
    \STATE $\mathit{planner} \gets \textsc{WaypointNavigator}(\text{target}{=}10\,\text{km/h})$
    \STATE $\mathit{planner}$.\textsc{SetWaypoints}($W$); \texttt{lane\_change\_forced} $\gets$ true
  \ENDIF
\ENDIF
\IF{reachedTarget or \textsc{EmvFullyPassed}()} \STATE \textsc{RestoreLaneChange}() \COMMENT{back to TM}
\ELSIF{$\mathit{planner}$ active} \STATE apply $\mathit{planner}$.\textsc{RunStep}() \ENDIF
\STATE \textbf{\texttt{npc\_behavior\_lvl}${=}2$: pull-over then braking hold}
\IF{no active planner} \STATE disable autopilot; $\mathit{planner} \gets \textsc{WaypointNavigator}(\text{target}{=}15\,\text{km/h})$ \ENDIF
\IF{$\mathit{planner}$ has no waypoints and alignment$\in\{L,R\}$}
  \STATE $\mathit{planner}$.\textsc{SetWaypoints}(\textsc{BuildPullOverWaypoints}(alignment)) \COMMENT{lane-edge, not SUMO copy}
\ENDIF
\IF{reachedTarget and not \textsc{HeadingAligned}() and alignment$\in\{L,R\}$}
  \STATE extend pull-over waypoints to finish lane alignment
\ENDIF
\IF{reachedTarget and \textsc{HeadingAligned}()}
  \STATE $\mathit{braking}\gets$ true; apply full brake $+$ hand-brake \COMMENT{stationary hold}
\ELSE
  \STATE apply $\mathit{planner}$.\textsc{RunStep}()
\ENDIF
\STATE \textbf{Release (both levels):} on \textsc{EmvFullyPassed}(), clear planner, re-enable autopilot, restore TM speed
\end{algorithmic}
\end{algorithm}

The EMV \texttt{gap\_escape} routine is the highest-priority recovery mode:
when the EMV stays below a low speed threshold long enough while a blocker is
detected (from the SUMO leader query or the navigator collision snapshot), the
controller repeatedly calls \texttt{build\_emv\_gap\_escape\_waypoints} to plan
a local path through available gap space and temporarily disables normal
collision avoidance. Its trigger and release constants, together with the other co-simulation control parameters, are given in Table~\ref{tab:thresholds}.

\begin{table}[htbp]
\centering
\small
\setlength{\tabcolsep}{4pt}
\renewcommand{\arraystretch}{1.12}
\caption{Control constants used by the co-simulation (from the controller
implementation).}
\label{tab:thresholds}
\resizebox{\columnwidth}{!}{%
\begin{tabular}{@{}llc@{}}
\toprule
Constant & Meaning & Value \\
\midrule
\multicolumn{3}{@{}l}{\emph{Co-simulation and control}}\\
Interaction region (in\,/\,out) & EMV-centered hysteresis radius & 40\,/\,50\,m \\
Leader query range & SUMO leader look-ahead (gap\_escape) & 80\,m \\
Lane-change cooldown & min.\ interval between forced-LC attempts & 2.0\,s \\
EMV-passed clearance & buffer for \textsc{EmvFullyPassed} & 1.0\,m \\
Heading-aligned tolerance & max.\ error for \textsc{HeadingAligned} & $8^{\circ}$ \\
\midrule
\multicolumn{3}{@{}l}{\emph{EMV} \texttt{gap\_escape}}\\
$v_{\mathrm{stop}}$ (\texttt{speed\_threshold\_mps}) & blocked-speed ceiling      & 0.5\,m/s \\
$t_{\mathrm{stop}}$ (\texttt{stop\_threshold\_s})    & sustained-stop before entry & 3.0\,s \\
$t_{\mathrm{grace}}$ (\texttt{blocker\_grace\_s})    & blocker-free time to exit   & 2.0\,s \\
\texttt{target\_speed}         & speed during recovery       & 20\,km/h \\
\bottomrule
\end{tabular}}
\end{table}

\subsubsection{Controller Target Speeds}
\label{sup:speeds}

Table~\ref{tab:speeds} lists the mode-dependent target speed the shared
\texttt{WaypointNavigator} enforces in each control mode. Values are in km/h
(the navigator converts internally to m/s). Nominal civilian motion is
governed by the CARLA Traffic Manager rather than a fixed navigator target.

\begin{table}[htbp]
\centering
\small
\setlength{\tabcolsep}{4pt}
\renewcommand{\arraystretch}{1.12}
\caption{Per-mode target speeds enforced by the shared waypoint navigator.}
\label{tab:speeds}
\begin{tabular}{@{}llc@{}}
\toprule
Agent & Control mode & Target (km/h) \\
\midrule
EMV      & Next-link route-following      & 50 \\
EMV      & Intent-override lane change    & 50 \\
EMV      & \texttt{copy\_sumo} corridor   & 30 \\
EMV      & \texttt{gap\_escape} recovery  & 20 \\
Civilian & Nominal (Traffic Manager)      & TM-governed \\
Civilian & L2 feasible-yield lane change  & 10 \\
Civilian & L3 pull-over (then braking hold) & 15 (then 0) \\
\bottomrule
\end{tabular}
\end{table}
Note that nominal civilian motion runs under the CARLA Traffic Manager at the
road speed limit (a $0\%$ Traffic-Manager speed difference). The L2
feasible-yield maneuver additionally sets the vehicle to $50\%$ below that limit
for the duration of the lane change.

\subsubsection{Data Formats and Coordinate Conventions}
\label{sup:formats}

Trajectory logs are stored as \texttt{trajectory.csv}, one row per actor per
logged frame, with columns \texttt{step}, \texttt{timestamp},
\texttt{actor\_id}, \texttt{type\_id}, \texttt{x}, \texttt{y}, \texttt{z},
\texttt{yaw}, \texttt{vx}, \texttt{vy}, \texttt{speed\_mps}, \texttt{is\_emv},
in the CARLA world frame ($x$ forward, $y$ right, $z$ up; yaw in degrees,
increasing clockwise, i.e.\ a left-handed frame) at 10~Hz; \texttt{is\_emv}
marks the ego emergency vehicle.

For 3D detection, \texttt{carla\_to\_kitti} exports KITTI-format point clouds
(\texttt{velodyne/}, float32 $x\,y\,z\,\text{intensity}$) and per-frame labels
(\texttt{label\_2/}). Points and boxes are placed in the EMV sensor frame by
rotating world coordinates by $-\psi_{\mathrm{EMV}}$ about the vertical axis,
with the LiDAR at $z=1.93$~m; objects beyond 50~m are dropped, matching the
nuScenes evaluation range. Since the
KITTI LiDAR frame is right-handed ($y$ points left) while CARLA is left-handed
($y$ points right), the exporter negates the lateral axis of the exported
boxes ($l_y\!\to\!-l_y$), matching the point cloud, which is already stored in
this convention. Since a reflection maps a heading $\theta\!\to\!-\theta$, the
sensor-frame heading is negated to stay consistent with the boxes. Omitting
this negation leaves predicted and label headings differing by
$\approx\!2\theta$; whereas keeping it, median orientation error is $\approx\!3^{\circ}$.
Each label line is
\texttt{type trunc occl alpha x1 y1 x2 y2 h w l lx ly lz heading}.

% ##########################################################################
% PART B --- SIREN-BENCH EVALUATION
% ##########################################################################
\subsection{SIREN-Bench Evaluation Details}
\label{sup:bench}

\subsubsection{Evaluation Metrics}
\label{sup:metrics}
% TODO(review): confirm the RMSE definition matches the NGSIM convention used
% by the code, and confirm whether minADE/minFDE use k=6 for all multi-modal
% models.

For a predicted trajectory $\hat{\mathbf{p}}_{1:T}$ and ground truth
$\mathbf{p}_{1:T}$ over $T$ future steps, with $\|\cdot\|$ the Euclidean norm,
we use the standard displacement metrics
\begin{align}
\mathrm{ADE} &= \frac{1}{T}\sum_{t=1}^{T}\lVert \hat{\mathbf{p}}_t-\mathbf{p}_t\rVert,
& \mathrm{FDE} &= \lVert \hat{\mathbf{p}}_T-\mathbf{p}_T\rVert .
\end{align}
For a multi-modal predictor emitting $k$ hypotheses
$\{\hat{\mathbf{p}}^{(i)}_{1:T}\}_{i=1}^{k}$, the min-of-$k$ variants report,
for each metric independently, the best of the $k$ hypotheses,
\begin{align}
\mathrm{minADE}_k &= \min_{i}\frac{1}{T}\sum_{t=1}^{T}
\lVert \hat{\mathbf{p}}^{(i)}_t-\mathbf{p}_t\rVert,\\
\mathrm{minFDE}_k &= \min_{i}\lVert \hat{\mathbf{p}}^{(i)}_T-\mathbf{p}_T\rVert .
\end{align}
The NGSIM-style RMSE reports the root-mean-square position error at the 5-s
horizon over the $N$ scored samples,
\begin{equation}
\mathrm{RMSE}@5\mathrm{s} = \sqrt{\frac{1}{N}\sum_{n=1}^{N}
\lVert \hat{\mathbf{p}}^{(n)}_{T}-\mathbf{p}^{(n)}_{T}\rVert^{2}} .
\end{equation}

For 3D detection, we follow the nuScenes protocol: a prediction matches a
ground-truth box when their bird's-eye-view center distance is below a
threshold $d\in\mathcal{D}=\{0.5,1,2,4\}$~m. With $\mathrm{AP}_d$ the
area under the precision--recall curve at threshold $d$,
\begin{equation}
\mathrm{mAP} = \frac{1}{|\mathcal{D}|}\sum_{d\in\mathcal{D}}\mathrm{AP}_d .
\end{equation}
The nuScenes detection score combines mAP with the five true-positive (TP)
error terms---translation (ATE), scale (ASE), orientation (AOE), velocity
(AVE), and attribute (AAE) errors---each averaged over matches and clipped to
$[0,1]$ as $1-\min(1,\mathrm{mTP})$:
\begin{equation}
\mathrm{NDS} = \frac{1}{10}\Big[5\,\mathrm{mAP} +
\sum_{\mathrm{mTP}\in\mathbb{TP}} \big(1-\min(1,\mathrm{mTP})\big)\Big],
\end{equation}
where $\mathbb{TP}=\{\mathrm{ATE,ASE,AOE,AVE,AAE}\}$.

\subsubsection{Evaluation Protocols and Model Setup}
\label{sup:proto}
% Fulfils (benchmark half of): main paper, Experimental Setup --- "Further
% implementation details appear in the supplementary material." All learned
% models use released pretrained checkpoints without fine-tuning on
% SIREN-Bench unless stated otherwise.

\paragraph{Trajectory prediction.}
All methods are scored on a common grid: a 5-s future at 0.2~s (25 steps),
one sample per (vehicle, frame) pair with a fully observed window, stride one
frame; every qualifying vehicle is scored, including the EMV. Observation
windows are fixed by each released checkpoint and differ by family: CSP,
STDAN, BAT, and the non-learned references (CV, IDM) use a 3-s history at
0.2~s (NGSIM convention), whereas EMP-M, EMP-D, and DeMo use a 5-s history at
0.1~s and natively emit a 6-s future, which we resample onto the common grid.
The templates yield 7{,}062 scored samples in total
(914/1{,}371/825/1{,}247/605/340/1{,}760 for S1--S7).
% TODO(impl): add per-model source dataset + checkpoint provenance, any
% coordinate-frame / unit conversion applied to SIREN inputs, and the
% resampling procedure for the Argoverse-native models. Confirm against the
% benchmarking_pipeline adapters.

\paragraph{3D object detection.}
Four LiDAR-based detectors pretrained on nuScenes are evaluated:
PointPillars, SECOND, VoxelNeXt, and TransFusion-L. Scoring follows the
nuScenes protocol of Sec.~\ref{sup:metrics}. Labeled vehicles all carry a
single \texttt{Car} label. Absolute values are not comparable to
published nuScenes numbers: the checkpoints run zero-shot on a denser 64-beam
sensor and a different city layout.
% TODO(impl): note the KITTI/nuScenes export path (carla_to_kitti) and any
% ego-frame / heading convention used when converting CARLA labels.

\subsubsection{Model and Checkpoint Provenance}
\label{sup:provenance}

Table~\ref{tab:provenance} lists the source and training domain of every
evaluated model; all learned models run zero-shot on SIREN-Bench from released
checkpoints without fine-tuning. The trajectory predictors split by training
domain---NGSIM highway car-following (CSP, STDAN, BAT) versus Argoverse~2 urban
driving (EMP, DeMo, RealMotion)---which the main paper flags as a confound
between families; CV and IDM are analytical references with no training data.
The four detectors use nuScenes-pretrained checkpoints. The five VLMs use the
exact released variants listed in the table, without SIREN-Bench fine-tuning;
their input preprocessing and prompting follow SeeUnsafe as described in
Sec.~\ref{sup:risk}.

\begin{table}[htbp]
\centering
\small
\setlength{\tabcolsep}{6pt}
\renewcommand{\arraystretch}{1.1}
\caption{Provenance of evaluated models. All learned models run zero-shot from
released pretrained checkpoints (no fine-tuning on SIREN-Bench).}
\label{tab:provenance}
\begin{tabular}{@{}>{\raggedright\arraybackslash}p{2.8cm} l
>{\raggedright\arraybackslash}p{3.2cm}
>{\raggedright\arraybackslash}p{2.3cm}@{}}
\toprule
Model & Task & Training data or checkpoint & Reference \\
\midrule
CV               & Traj. & --- (analytical)   & --- \\
IDM              & Traj. & --- (car-following) & --- \\
CSP              & Traj. & NGSIM              & \citep{deo2018csp} \\
STDAN            & Traj. & NGSIM              & \citep{chen2022stdan} \\
BAT              & Traj. & NGSIM              & \citep{liao2024bat} \\
EMP-M / EMP-D    & Traj. & Argoverse~2        & \citep{prutsch2024emp} \\
DeMo             & Traj. & Argoverse~2        & \citep{zhang_demo_2024} \\
RealMotion       & Traj. & Argoverse~2        & \citep{song2024realmotion} \\
\midrule
PointPillars     & Det.  & nuScenes           & \citep{lang2019pointpillars} \\
SECOND           & Det.  & nuScenes           & \citep{yan2018second} \\
VoxelNeXt        & Det.  & nuScenes           & \citep{chen2023voxelnext} \\
TransFusion-L    & Det.  & nuScenes           & \citep{bai2022transfusion} \\
\midrule
Blaifa-InternVL  & Risk  & Blaifa-InternVL3.5-8B & \citep{wang2025internvl35} \\
Gemma 3          & Risk  & Gemma3-12B          & \citep{gemma_2025} \\
LLaVA-Llama3     & Risk  & LLaVA-Llama3-8B     & \citep{2023xtuner} \\
MiniCPM-V 4.5    & Risk  & MiniCPM-V-4.5-8B    & \citep{yu_minicpm-v_2025} \\
Qwen 3.5         & Risk  & Qwen3.5-9B          & \citep{qwen2026qwen35} \\
\bottomrule
\end{tabular}
\end{table}

\subsubsection{Extended Trajectory-Prediction Metrics}
\label{sup:traj}
% Fulfils: main paper, Trajectory-prediction setup --- "We report ADE and FDE
% here; minADE_6, minFDE_6, and 5-s RMSE are in the supplementary material."
% Numbers lifted from benchmarking_pipeline/benchmark_tables.tex (7-scene eval).

Table~\ref{tab:rmse} reports NGSIM-style 5-s RMSE per template for all nine
predictors, and Table~\ref{tab:minmetrics} reports multimodal best-of-6
minADE$_6$/minFDE$_6$ for the four map-conditioned models; both use the common
5-s / 0.2-s grid of Table~1 in the main paper. Being a root-mean-square, RMSE
is dominated by the largest per-step errors and exceeds the mean-of-distance
ADE for every model.

\begin{table}[htbp]
\centering
\caption{NGSIM-style RMSE at the 5-s horizon for each interaction template.
Lower is better; values are in meters. All methods use the same evaluation
grid as Table~\ref{tab:prediction}. $^{\ddagger}$ denotes Argoverse~2
map-conditioned models evaluated zero-shot.}
\label{tab:rmse}

\small
\setlength{\tabcolsep}{4pt}
\renewcommand{\arraystretch}{1.12}

\begin{adjustbox}{max width=\linewidth}
\begin{tabular}{@{}lcccccccc@{}}
\toprule
& \multicolumn{2}{c}{Clearance}
& \multicolumn{4}{c}{Traffic-control traversal}
& \multicolumn{1}{c}{Road-space}
& \\
\cmidrule(lr){2-3}
\cmidrule(lr){4-7}
\cmidrule(lr){8-8}
Model & S1 & S2 & S3 & S4 & S5 & S6 & S7 & Mean \\
\midrule
CV      & 5.61  & 5.26  & 4.64  & 3.56  & 6.49  & 8.55  & 4.82  & 5.56  \\
IDM     & 7.06  & 6.40  & 5.16  & 3.88  & 8.01  & 9.20  & 5.72  & 6.49  \\
CSP     & 7.52  & 10.07 & 8.67  & 7.27  & 8.76  & 11.16 & 5.84  & 8.47  \\
STDAN   & 10.38 & 16.71 & 11.24 & 9.47  & 12.39 & 21.15 & 10.38 & 13.10 \\
BAT     & 18.35 & 17.76 & 20.20 & 14.35 & 9.70  & 14.55 & 8.31  & 14.75 \\
\midrule
EMP-M$^{\ddagger}$
        & 6.39  & 6.34  & 10.54 & 9.24  & 12.19 & 11.05 & 9.14  & 9.27  \\
EMP-D$^{\ddagger}$
        & 5.52  & 4.81  & 9.56  & 7.05  & 7.82  & 12.11 & 7.51  & 7.77  \\
DeMo$^{\ddagger}$
        & 5.35  & 5.62  & 9.22  & 6.66  & 9.80  & 12.31 & 5.89  & 7.84  \\
DeMo+RealMotion$^{\ddagger}$
        & 5.97  & 6.01  & 10.86 & 10.10 & 8.95  & 8.37  & 8.16  & 8.34  \\
\bottomrule
\end{tabular}
\end{adjustbox}
\end{table}

\begin{table}[htbp]
\centering
\caption{Multimodal best-of-six minADE$_6$/minFDE$_6$ at the 5-s horizon
for the four map-conditioned models. Lower is better; values are in meters.
The anchors and NGSIM-based models are single-mode, so their
minADE$_6$/minFDE$_6$ equal the ADE/FDE values reported in
Table~\ref{tab:prediction}.}
\label{tab:minmetrics}

\small
\setlength{\tabcolsep}{4pt}
\renewcommand{\arraystretch}{1.12}

\begin{adjustbox}{max width=\linewidth}
\begin{tabular}{@{}lcccccccc@{}}
\toprule
& \multicolumn{2}{c}{Clearance}
& \multicolumn{4}{c}{Traffic-control traversal}
& \multicolumn{1}{c}{Road-space}
& \\
\cmidrule(lr){2-3}
\cmidrule(lr){4-7}
\cmidrule(lr){8-8}
Model & S1 & S2 & S3 & S4 & S5 & S6 & S7 & Mean \\
\midrule
\multicolumn{9}{@{}l}{\emph{minADE}$_6$} \\
EMP-M
& 0.58 & 0.52 & 0.89 & 0.66 & 1.06 & 0.93 & 0.58 & 0.74 \\
EMP-D
& 0.58 & 0.48 & 0.81 & 0.52 & 0.81 & 0.80 & 0.37 & 0.62 \\
DeMo
& 0.53 & 0.47 & 0.67 & 0.39 & 0.90 & 0.99 & 0.32 & \textbf{0.61} \\
DeMo+RealMotion
& 0.58 & 0.41 & 0.93 & 0.77 & 0.82 & 0.80 & 0.50 & 0.69 \\
\midrule
\multicolumn{9}{@{}l}{\emph{minFDE}$_6$} \\
EMP-M
& 1.30 & 1.26 & 2.72 & 2.20 & 2.71 & 2.34 & 2.25 & 2.11 \\
EMP-D
& 1.21 & 1.07 & 2.44 & 1.58 & 1.96 & 1.96 & 1.30 & \textbf{1.65} \\
DeMo
& 1.12 & 1.24 & 2.18 & 1.34 & 2.31 & 2.58 & 1.60 & 1.77 \\
DeMo+RealMotion
& 1.28 & 1.05 & 2.72 & 2.34 & 2.00 & 1.95 & 1.59 & 1.85 \\
\bottomrule
\end{tabular}
\end{adjustbox}
\end{table}

\subsubsection{Detection True-Positive Error Breakdown}
\label{sup:det}
% Fulfils: main paper, 3D object detection --- "The detailed true-positive
% error breakdown is deferred to the supplement." Content migrated from the
% original supplementary_appendix.tex fragment.

Table~\ref{tab:t2-tp} reports the nuScenes true-positive error terms that
enter NDS, averaged over the seven templates. Translation and scale errors
are small and consistent across detectors (ATE $0.55$--$0.57$~m, ASE
$0.20$--$0.22$), confirming that the detectors localize and size vehicles
correctly; orientation error separates the architectures as expected.

The velocity and attribute terms behave differently. AVE exceeds
$3.1$~m/s per detector, far above typical nuScenes values. We
checked this against the ground-truth velocity frame and against
alternative sign and axis conventions, and the reported convention fits best. Hence, we treat the magnitude as a genuine cross-domain effect rather
than a unit or frame error: the detectors are asked to regress velocity
for traffic whose speed profile is different from the nuScenes training distribution; for example, stop-and-go queues interrupted by a fast EMV. Similar patterns are observed under AAE,
as CARLA supplies no attribute labels, and the moving/stopped
attribute is derived from speed. Both terms clip toward the low end of
NDS, which is why we report mAP as the headline detection metric and NDS
as secondary.

\begin{table}[htbp]
\centering
\caption{True-positive error metrics for 3D object detection, averaged
over the seven templates. Lower is better; units: ATE\,m, AOE\,rad,
and AVE\,m/s.}
\label{tab:t2-tp}

\small
\setlength{\tabcolsep}{4pt}
\renewcommand{\arraystretch}{1.12}

\begin{adjustbox}{max width=\linewidth}
\begin{tabular}{@{}lccccc@{}}
\toprule
Model
& ATE $\downarrow$
& ASE $\downarrow$
& AOE $\downarrow$
& AVE $\downarrow$
& AAE $\downarrow$ \\
\midrule
PointPillars  & 0.553 & 0.215 & 0.396 & 3.165 & 0.459 \\
SECOND        & 0.565 & 0.214 & 0.207 & 3.196 & 0.514 \\
VoxelNeXt     & 0.562 & 0.216 & 0.313 & 3.404 & 0.576 \\
TransFusion-L & 0.561 & 0.196 & 0.330 & 3.223 & 0.461 \\
\bottomrule
\end{tabular}
\end{adjustbox}
\end{table}

\subsubsection{Risk Understanding}
\label{sup:risk}

The risk-understanding evaluation uses the preprocessing and prompting
components of \textit{SeeUnsafe}~\cite{zhang2025seeunsafe} unchanged,
including its segmentation- and visual-grounding-based visual prompts and
accompanying text prompt. We apply this pipeline to the SIREN-Bench videos
and reference labels; implementation details follow SeeUnsafe and are not
repeated here.
The task treats risk classification as a behavioral measure of scene-level
risk understanding: inferring the class requires interpreting the agents and
their interactions, but the evaluation scores the resulting classification
rather than the VLM's intermediate reasoning process.
The source data comprise 105 front-camera videos, with 15 videos for each of
the seven interaction templates and 190 frames per video, yielding 19{,}950
frames in total. We evaluate
the five vision-language models reported in the main paper:
Blaifa-InternVL3.5-8B, Gemma3-12B, LLaVA-Llama3-8B,
MiniCPM-V-4.5-8B, and Qwen3.5-9B.

Each MP4 video produces two intermediate predictions assigned to one of three
risk classes: Normal, Near-Miss, or Collision. Following SeeUnsafe, the more
severe prediction is selected using the ordering Normal $<$ Near-Miss $<$
Collision, yielding one final prediction per video. The evaluation therefore
contains 105 video-level predictions per model. Reference labels
are derived from the recorded EMV--civilian interactions and collision events.
Evaluation is based on
overall classification accuracy and class-specific precision and F1.
The class-specific results are reported in
Table~\ref{tab:risk_classification} of the main paper.

% ============================================================
%  Acknowledgments
% ============================================================
\section*{Acknowledgements}
This work was supported by the Small World Lab (SWL) at Rochester Institute of Technology. Haoxin Leng and Tao Li are supported by CityU internal grant No. 9610792. The contents of this paper only reflect the views of the authors who are responsible for the facts and do not represent any official views of any sponsoring organizations or agencies.

\section*{Author Contributions}
\noindent\textbf{Yicheng Zhu}: Pipeline methodology, software, validation, and writing.\par
\noindent\textbf{Tianmu Zhao}: Behavior-model methodology, software, and writing---review \& editing.\par
\noindent\textbf{Haoxin Leng}: Data collection and evaluation pipeline.\par
\noindent\textbf{Fan Zuo}: Conceptualization and writing---review \& editing.\par
\noindent\textbf{Tao Li}: Conceptualization and writing---review \& editing.\par
\noindent\textbf{Zilin Bian}: Conceptualization, methodology, supervision, and writing---review \& editing.

% ============================================================
%  References
% ============================================================
\bibliographystyle{unsrtnat}
\bibliography{references,aaai2027}

\end{document}